%% file: 0-main.tex
\documentclass[runningheads]{llncs}
\usepackage[subtle, tracking=normal]{savetrees}
\usepackage[T1]{fontenc}
\usepackage{orcidlink}
\usepackage{amsmath}
\usepackage{graphicx}
\usepackage{booktabs}
\usepackage{algpseudocode}
\usepackage{algorithm}
\usepackage{tabularx}
\usepackage{setspace}
\usepackage{xcolor}
\usepackage{orcidlink} % For ORCID icons
\usepackage{amssymb} % already loaded
\usepackage{ulem}
\usepackage[inline]{enumitem}
\newcommand{\xmark}{\textcolor{red!70!black}{\(\times\)}} % darker red
\newcommand{\cmark}{\textcolor{green!75!black}{\(\checkmark\)}} % darker green

\begin{document}
\title{SCOPE: Sequential Causal Optimization\\ of Process Interventions}
\titlerunning{SCOPE}

\author{
Jakob De~Moor\inst{1}\orcidlink{0009-0006-4788-5346} \and
Hans Weytjens\inst{1,2}\orcidlink{0000-0003-4985-0367} \and \\
Johannes De~Smedt\inst{1}\orcidlink{0000-0003-0389-0275} \and
Jochen De~Weerdt\inst{1}\orcidlink{0000-0001-6151-0504}
\thanks{This work was supported by the Research Foundation - Flanders (11A6J25N and G039923N), and Internal Funds KU Leuven (C14/23/031).}
}

\institute{
Research Centre for Information Systems Engineering (LIRIS),\\ KU Leuven, Leuven, Belgium\\
\email{\{jakob.demoor, hans.weytjens, johannes.desmedt\}@kuleuven.be}
\and
School of Computation, Information and Technology, \\Technical University of Munich (TUM), Munich, Germany\\
\email{hans.weytjens@tum.de}
}

\authorrunning{De Moor et al.}

% \makeatletter
% \renewcommand{\maketitle}{%
%   \begin{center}
%     {\LARGE\bfseries \@title \par}
%     \vskip 1em
%   \end{center}
% }
% \makeatother

% \author{}
% \authorrunning{}
% \institute{}
\maketitle              % typeset the header of the contribution
\vspace{-10pt}
\begin{abstract}
Prescriptive Process Monitoring (PresPM) recommends interventions during running business processes to optimize key performance indicators (KPIs).
In realistic settings, interventions are rarely isolated: organizations need to align sequences of interventions to jointly steer the outcome of a case. Existing PresPM approaches only partially address this challenge. Many focus on a single intervention decision, while others treat multiple interventions independently, ignoring how they interact over time. Methods that do address these dependencies depend either on simulation or data augmentation to approximate the process to train a Reinforcement Learning (RL) agent, which may create a reality gap and introduce bias. We introduce \textit{SCOPE} (\underline{S}equential \underline{C}ausal \underline{O}ptimization of \underline{P}rocess Interv\underline{e}ntions), a PresPM approach that learns aligned sequential intervention recommendations. \textit{SCOPE} employs backward induction to estimate the effect of each candidate intervention action, propagating its impact from the final decision point back to the first. By leveraging causal learners, our method can utilize observational data directly, unlike methods that require constructing process approximations for RL. Experiments on both an existing synthetic dataset and a new semi-synthetic dataset show that \textit{SCOPE} consistently outperforms state-of-the-art PresPM techniques in optimizing the KPI. The novel semi-synthetic setup, based on a real-life event log, is provided as a reusable benchmark for future work on sequential PresPM.
\vspace{-5pt}
\keywords{Prescriptive Process Monitoring \and Sequential Decision-Making \and Causal Learning}
\vspace{-10pt}
\end{abstract}
\input{1-introduction}
\input{2-background}
\input{3-methodology}
\input{4-experiments_discussion}
\input{5-conclusion}
\bibliographystyle{splncs04}
\bibliography{references}

\end{document}

%% file: 1-introduction.tex
\section{Introduction}\label{sec:introduction}
\vspace{-9pt}
PresPM uses machine learning to provide case-specific recommendations at different decision points during the execution of business processes. These recommendations concern interventions, such as managerial escalations or customer communications, that aim to improve KPIs, for example, throughput time or cost efficiency. PresPM holds the potential to move organizations from merely describing and predicting process behavior towards actively steering process executions~\cite{bozorgi2023CI}. In this paper, we use the terms \textit{intervention} for any controllable action taken on an ongoing case, and \textit{intervention recommendation} for the action suggested by a PresPM method at a \textit{decision point}.

In many processes, intervention decisions are not isolated. Typically, a process contains multiple, interdependent decision points that jointly determine the outcome of a case. The effect of an earlier (or later) intervention depends on which interventions are applied later (or earlier) in the same case. Optimizing intervention decisions one by one is therefore insufficient, because actions that look beneficial locally can undermine overall KPI performance.  
For example, in a marketing process aimed at maximizing revenue, the effectiveness of offering a client a discount may depend on an earlier choice to send a promotional email and the client’s response to that email. This sequence of decisions together shapes the revenue. Optimizing each decision independently without considering the future might make it look like sending a promotional email is only moderately valuable for some clients, even though if you follow it with a discount, it could be highly effective. Methods for PresPM thus need to reason over sequences of decisions and their combined effect on the final KPI.

Most existing PresPM approaches do not yet offer such sequential support. 
First, many methods address only a single intervention scenario, even when multiple opportunities to intervene exist~\cite{bozorgiLearningWhenToTreat,bozorgiCostAwareCycleTime,bozorgi2023CI,ShoushResRL,shoushconformal,shoushResCI,shoushwhitebox}. These approaches may improve performance for that single intervention scenario, but they do not coordinate multiple decisions over the full case. %,whereas business processes often involve multiple intervention decisions that need to be coordinated.
Second, some approaches handle sequential decisions but optimize each decision point in isolation. They focus on the immediate effect of the next intervention, for instance, on the remaining processing time, without explicitly aligning interventions across decision points. This can still lead to suboptimal end-to-end outcomes with respect to the final KPI~\cite{leoni2020}.
Third, other methods for sequential intervention recommendations rely on process approximations, such as Markov decision processes (MDP) or data augmentation~\cite{abbasi2025,branchi2022}, to train an RL agent, and the resulting policies inherit any misspecification in these approximations, also known as the reality gap, which may lead to biased or underperforming recommendations in practice~\cite{realitygap}.

To address these limitations, this paper makes two key contributions:
\begin{enumerate}[topsep=0pt, partopsep=0pt, itemsep=0pt, parsep=0pt]
    \item We propose \textit{SCOPE}, a PresPM approach that combines causal learners with backward induction to learn causally grounded sequential intervention policies that are aligned across multiple decision points. For each decision point, a causal model estimates the effect of alternative interventions on the target KPI, given the observed process execution history. During training, backward induction then propagates the impact of later intervention decisions back to earlier ones by starting from the final decision point and recursively deriving the optimal action and its expected outcome at each preceding decision point. Operating directly on observational event logs, \textit{SCOPE} uses causal learning to be able to identify actions rarely chosen historically but likely to enhance the target KPI, without requiring process-specific simulators or log augmentation.
    \item We provide an empirical evaluation on existing synthetic data and a new semi-synthetic setup based on a real-life event log. Our code and the novel semi-synthetic benchmark are made publicly available as a reusable resource for sequential PresPM in our \href{https://github.com/JakobDeMoorKULstudent/SCOPE}{GitHub repository}. The results show that \textit{SCOPE} outperforms the considered baselines across a broad set of experimental settings, highlighting the importance of combining causal learning with backward induction for KPI optimization.
\end{enumerate}

The paper is organized as follows: Section \ref{sec:background} reviews background and related work, Section \ref{sec:methodology} describes the methodology, Section \ref{sec:experiments_discussion} covers experiments, results, and discussion, and Section \ref{sec:conclusion} concludes the paper.
\vspace{-10pt}

%% file: 2-background.tex
\section{Background}\label{sec:background}
\vspace{-11pt}
\subsection{General}
\vspace{-8pt}
% \vspace{-10pt}
Current PresPM approaches are generally based on two streams of research. 
The first is Causal Inference (CI), which aims to estimate the effects of potential interventions from observational data and use these estimates to guide decisions. This is achieved through causal learners, which are model setups designed to use observational data to predict outcomes under an action and/or causal effects of choosing one action over the other. For example, an S-learner fits a single model that predicts the outcome using the intervention action as an input in addition to other features. PresPM approaches typically adapt causal learners to the process context by aggregating event data for use with standard CI setups~\cite{bozorgi2023CI}, or by leveraging sequential models like LSTMs~\cite{weytjens2023}.
\footnote{CI typically follows causal discovery, which identifies causal relationships, while CI estimates their magnitude~\cite{weytjens2023}. Most PresPM studies that optimize interventions (implicitly) adopt a known standard causal structure with time-varying confounders, interventions and an outcome.}
The second stream is RL-based, where an agent interacts with an environment that represents/approximates the real process, observes states, selects actions, and receives rewards, with the goal of learning a policy that maximizes cumulative rewards~\cite{ShoushResRL}. 

Similar to CI, PresPM typically relies on offline observational event logs, rather than data from controlled experiments. This approach avoids the cost and risk associated with randomized experiments, such as randomized controlled trials (RCTs) or an online RL agent exploring a real-life environment, which are often not feasible (e.g., when testing a loan assignment strategy).
However, observational data introduces challenges. Because the data reflects an existing \textit{historical decision policy} (e.g., a bank’s current loan strategy), treatment/intervention assignment is not random as it would be in controlled trials. This complicates the estimation of optimal actions due to \textit{confounding} factors affecting both intervention assignment and outcomes~\cite{igcnet}. For example, if a bank offers optional financial literacy workshops (intervention) to improve repayment rates (KPI), individuals who choose to attend may already have stronger financial habits (confounder), making it difficult to isolate the workshop’s true effect. 
% Another challenge is model evaluation using observational data. We cannot always directly verify whether the model’s recommendation would have led to a better outcome for a test case. We only observe what actually happened under the historical decision policy, not what would have happened under a different intervention. This missing counterfactual is known as the fundamental problem of CI~\cite{holland1986}.
\vspace{-11pt}

\subsection{Related Work}
\vspace{-5pt}
This subsection reviews existing PresPM approaches, highlighting their key features and limitations. Table~\ref{tab:positioning} summarizes how \textit{SCOPE} compares to these methods, with the discussion below detailing the distinguishing dimensions.
\vspace{-10pt}

\subsubsection{Single-Interventional Approaches.}
Most existing PresPM methodologies focus on the impact of a single intervention scenario. For example, Bozorgi et al.~\cite{bozorgi2023CI} use causal effect estimation to identify which cases would benefit from one specific intervention decision, applying CI techniques after encoding event data into a suitable format. In their work, an example of such an intervention would be skipping a particular activity. Similarly, Shoush et al.~\cite{shoushwhitebox} propose a white-box framework that integrates predictive, causal, and survival models to recommend single intervention decisions. In their case, an example intervention would be offering either multiple loan options to a client or just one. Yet, real-world business processes often require multiple (interdependent) interventions rather than a single one, as they consist of interconnected events. In Table~\ref{tab:positioning}, the first column shows whether a method is multi-interventional or not.
% For instance, in a loan application process, a bank may want to optimize the sequential intervention decisions of requesting additional documents, performing a credit check, scheduling an interview, and escalating the case. In Table~\ref{tab:positioning}, the first column shows whether a method is multi-interventional or not.
\vspace{-10pt}

\subsubsection{Multi-Interventional Approaches.}
Some PresPM approaches handle multiple interventions by recommending the next best activity to optimize a KPI. Typically, predictive models estimate KPIs for possible actions at each decision point; for instance, de Leoni et al.~\cite{leoni2020} use a transition-system to identify valid next activities and select the one maximizing the predicted KPI. However, optimizing only the immediate action without considering subsequent decision points can yield suboptimal results. Consider again the revenue-maximizing marketing process with two consecutive decisions: sending a promotional email (decision point 1), then offering a discount (decision point 2). Assessing actions at decision point 1 without accounting for decision point 2 yields predictions that are \textit{averages over the historical distribution of actions taken at decision point 2}. Consequently, the model might undervalue an email that is only highly effective when followed by a discount, ignoring the cumulative effect of action sequences. Making optimal decisions requires aligning action sequences across all decision points. Column 2 in Table~\ref{tab:positioning} categorizes whether methods align interventions across multiple decision points or optimize at a single point, as in~\cite{leoni2020}.

A different line of work that addresses sequential interventions is presented by Abbasi et al.~\cite{abbasi2025} and Branchi et al~\cite{branchi2022}. Both aim to recommend the next best activity while explicitly accounting for all possible sequences of future actions. Branchi et al. achieve this by (1) applying KMeans clustering to group process prefixes in an event log; (2) defining an MDP where states are represented by the cluster label and previous activity, rewards are defined as the average reward observed for that state in the log, and transition probabilities are obtained by replaying the log; and (3) applying offline RL (Q-learning specifically) to derive a next-best-action policy~\cite{branchi2022}. Because Q-learning optimizes cumulative return over a full case, the resulting recommendations are aligned across decision points.\footnote{Branchi et al. \cite{branchi2022} extend their earlier work \cite{branchilearningtoact} by removing process-specific assumptions and thus improving generalizability, which is why we focus on it here.} Abbasi et al. take a slightly different route by also applying offline RL, but combined with data augmentation~\cite{abbasi2025}. They modify case timestamps and remove non-critical activities, arguing that these transformations help diversify the training data. 
However, both approaches rely on approximations of the real process through simulation (in the form of an MDP) or data augmentation. These approximations may lead to policies that underperform in practice, a phenomenon known as the reality gap~\cite{realitygap}. For example, in Branchi et al., important prefix information may be lost when aggregating cases prior to clustering, and KMeans may struggle with high-dimensional or categorical data. In Abbasi et al., augmentation is constrained by manually crafted rules: timestamps are only varied within 10\%, and three business rules (precedence, co-occurrence, conflict) guide modifications. Some processes may require different temporal bounds or a far richer set of constraints. In such situations, the learned policy may underperform. In Table~\ref{tab:positioning}, the last column indicates whether a method requires an approximation (e.g., a simulation or augmentation) of the process, or can directly use the data as-is.

We also note Weinzierl et al.~\cite{weinzierl2020}, whose approach predicts the most likely future suffix and identifies similar historical cases to select actions with favorable KPIs. By only exploring the neighborhood of the (historically) most likely suffix, this method depends heavily on the historical decision policy and risks missing superior but rarely observed actions. We therefore do not pursue this method further.
% \footnote{\label{fn:weinzierl}We also note the approach by Weinzierl et al.~\cite{weinzierl2020}, which predicts the most likely future suffix and identifies similar historical cases to select actions with favorable KPIs. This method heavily relies on behaviors seen under the historical decision policy that generated the training data, as it only considers the neighborhood of the most likely suffix, which may miss better, but rarely observed actions. Therefore, we do not consider this method further in this paper.}.

In summary, Table~\ref{tab:positioning} highlights a gap in current methods that \textit{SCOPE} aims to address.
% \vspace{-10pt}

\input{tables/positioning}
\vspace{-5pt}

\subsubsection{Sequential Decision-Making \& CI.}
Sequential decision-making approaches are also used in CI, commonly studied under the framework of Dynamic Treatment Regimes (DTR). DTR methods aim to identify optimal treatment policies to maximize long-term outcomes, typically in medical settings.
These methods rely on sequential optimization (similar to RL) and causal methods. For instance, in~\cite{schulte2014}, the authors discuss Q-learning, essentially a form of backward induction, to determine the optimal treatment regime. Another approach models the DTR objective as a classification problem~\cite{cclearning}, using causal learners and a classifier to recommend the best treatment at each step.
We argue that the sequential PresPM problem is related to DTRs and that DTR methods provide a promising foundation to address the gap identified in Table~\ref{tab:positioning}. However, mapping PresPM to the DTR framework is not straightforward. DTR typically assumes a fixed number of decision points (often medical stages), while PresPM operates on event-log prefixes from business processes with variable control flow. Case evolution depends on event-, case-, and activity-level variables, and may include loops, parallelism, and multiple branching structures. Consequently, the number of decision points varies across cases, and decision histories/prefixes can be highly heterogeneous at a decision point.
We therefore build on the DTR framework, but adapt and configure it to the sequential PresPM setting to identify intervention recommendations that maximize process KPIs in business settings.
% \footnote{DTRs are closely related to causal RL, which examines sequential decision-making under explicit causal assumptions within the RL framework~\cite{causalrl}. In offline, finite-horizon settings—such as business processes—DTRs can be seen as a statistically grounded form of causal RL.} 
\vspace{-8pt}

%% file: tables/positioning.tex
\begin{table}[ht!]
\centering
\caption{Comparison of PresPM methods for sequential interventions.}\label{tab:positioning}
\tiny
\renewcommand{\arraystretch}{1.1} % 1.5 = 50% taller rows
\setlength{\tabcolsep}{5pt} % increase column spacing
\begin{tabularx}{1\textwidth}{l X X X} % X = auto-width column
% \toprule
%  & \textbf{Multi-interventional} & \textbf{Consideration of actions rarely observed in data} & \textbf{Aligned prescriptions across stages } & \textbf{No process simulation or augmentation Required} \\
% \midrule
\toprule
 & \textbf{Multi-interventional} & \textbf{Aligned interventions across decision points} & \textbf{No process approximations required} \\
\midrule
% {\tiny \textit{Description}} & {\tiny Indicates whether the method handles sequential PresPM problems.} & {\tiny Specifies whether prescriptions in one stage account for those in other stages.} & {\tiny Indicates whether the method can be trained without simulated or augmented data approximating the real process.} \\
% {\tiny \textit{Why It Matters}} & {\tiny Many business processes require a series of different decisions, where each step shapes what happens next.} & {\tiny Methods limited to frequent actions under the historical decision policy cannot uncover potentially better but rarely chosen decisions.} & {\tiny A prescription optimal within one stage may be suboptimal across stages.} & {\tiny Simulated or augmented data approximations may misrepresent real processes, leading to suboptimal policies.} \\
{\tiny \textit{Why It Matters}} & {\tiny Many business processes require a series of different intervention decisions, where each step shapes what happens next.} & {\tiny An intervention recommendation optimal at one decision point may be suboptimal across decision points.} & {\tiny Using an MDP or augmented data to approximate processes for training may misrepresent true behavior, leading to suboptimal policies.} \\
% \midrule
\textbf{Method} &  &  &  \\
\midrule
\cite{bozorgiLearningWhenToTreat,ShoushResRL} & \parbox[c]{\linewidth}{\centering \xmark} & \parbox[c]{\linewidth}{\centering \xmark} & \parbox[c]{\linewidth}{\centering \xmark} \\
\midrule
\cite{bozorgiCostAwareCycleTime,bozorgi2023CI,shoushconformal,shoushResCI,shoushwhitebox} & \parbox[c]{\linewidth}{\centering \xmark} & \parbox[c]{\linewidth}{\centering \xmark} & \parbox[c]{\linewidth}{\centering \cmark}\\
\midrule
\cite{leoni2020} & \parbox[c]{\linewidth}{\centering \cmark} & \parbox[c]{\linewidth}{\centering \xmark} & \parbox[c]{\linewidth}{\centering \cmark}\\
\midrule
% \cite{weinzierl2020} & \parbox[c]{\linewidth}{\centering \cmark} & \parbox[c]{\linewidth}{\centering \xmark} &  \parbox[c]{\linewidth}{\centering \cmark} & \parbox[c]{\linewidth}{\centering \cmark}\\
% \midrule
\cite{abbasi2025} & \parbox[c]{\linewidth}{\centering \cmark} & \parbox[c]{\linewidth}{\centering \cmark} & \parbox[c]{\linewidth}{\centering \xmark} \\
\midrule
\cite{branchi2022} & \parbox[c]{\linewidth}{\centering \cmark} & \parbox[c]{\linewidth}{\centering \cmark} & \parbox[c]{\linewidth}{\centering \xmark} \\
\midrule
\textit{SCOPE} (ours) & \parbox[c]{\linewidth}{\centering \cmark} & \parbox[c]{\linewidth}{\centering \cmark} & \parbox[c]{\linewidth}{\centering \cmark}\\
\bottomrule
\end{tabularx}
\vspace{-15pt}
\end{table}

%% file: 3-methodology.tex
\section{Methodology}\label{sec:methodology}
\vspace{-5pt}
This section introduces definitions, notation, and a detailed explanation of \textit{SCOPE}.
\vspace{-20pt}

\subsection{Preliminaries}
% \vspace{-5pt}
\begin{definition}[Event, Event Log, Trace, Prefix]
An \textsc{event} is a tuple $e = (c, o, t, d, s)$, comprising a case identifier $c \in C$, activity label $o \in O$, timestamp $t \in \mathbb{R}^+$, optional event-specific attributes $d = (d_1, \dots, d_{m_d})$, and static case attributes $s = (s_1, \dots, s_{m_s})$.
An \textsc{event log} $L = \{ e_i\}_{i=1}^N$ is a collection of $N$ observed events.
A \textsc{trace} $\sigma=\langle e_1, \dots, e_{|\sigma|} \rangle$ is the time-ordered sequence of all events for a single case. A \textsc{prefix} consists of the first $l$ events of a trace.
\end{definition}
\vspace{-9pt}

\begin{definition}[Decision point, Intervention action, Policy, Outcome]
A \textsc{decision point} $k \in \{1, \dots, K\}$ is a prespecified or discovered point in the process where an intervention may occur. For a case $c$, let $\mathcal{K}^{(c)} \subseteq \{1, \dots, K\}$ denote the specific decision points it actually visits, noting that some may be skipped due to variable control flow.
% \footnote{\label{fn:skip}Due to variable control flow in business processes, historical cases may skip one or more decision points.}
The history available at $k$ is represented by the prefix $\sigma_k$, encompassing all events (and interventions) preceding $k$, and its length depends on the control flow it followed.
An \textsc{intervention action} $a_k \in A_k$ is chosen from a feasible space $A_k$, which maps to an existing attribute in the event log (i.e., activity label or any other event-specific attribute from $s$).\footnote{The action space can be predefined or discovered, e.g., as in~\cite{leoni2020}.} The historically observed action is denoted $a^{obs}_k$.
A \textsc{policy} $\pi = \{ \pi_1, \ldots, \pi_K \}$ is a set of decision rules mapping a prefix to an action: $\pi_k: \sigma_{k} \to a_k$.
The \textsc{outcome} $y$ is the trace's final observed KPI. Consequently, an event log $L$ can be encoded into the dataset:
$ \mathcal{D} = \bigcup_{c \in C} \bigcup_{k \in \mathcal{K}^{(c)}} \Big\{ \big(\sigma_{k}^{(c)}, a_k^{obs,(c)}, y^{(c)}\big) \Big\} $.
\end{definition}
\vspace{-15pt}

\paragraph{Objective.}
The goal is to find the optimal policy $\pi^{opt} = \{ \pi^{opt}_1, \ldots, \pi^{opt}_K \}$, given by $\pi^{opt} = \arg\max_{\pi} \mathbb{E} \big[ y({\pi
})\big]$, where $y(\pi)$ is the potential outcome under the policy $\pi$~\cite{POframework}.
\vspace{-5pt}

\subsection{\textit{SCOPE}}
\vspace{-5pt}
Below, we explain how \textit{SCOPE} maximizes a KPI (use $min$ instead of $max$ for minimization). First, we outline backward induction theory. Next, we describe how causal learners are integrated. Finally, we present the full \textit{SCOPE} algorithm.
% \vspace{-5pt}

\subsubsection{Backward Induction.}
We now explain how backward induction (theoretically) identifies the optimal sequential intervention policy, adapted here to the business process context with DTR histories as prefixes with variable control-flow. Following DTR and PresPM literature, we begin with three standard assumptions: sequential ignorability, the stable unit treatment value assumption (SUTVA), and positivity.\footnote{Sequential ignorability: no unmeasured confounders given a prefix; SUTVA: intervening on a case does not affect another case’s outcome and interventions have well-defined versions; positivity: given a prefix, every feasible action has a nonzero chance of occurring historically.} Under these assumptions, the optimal policy $\pi^{opt}$ is identifiable from observational data through backward induction, meaning it \textit{can}, in principle, be obtained (see \cite{schulte2014} for a detailed discussion on these assumptions and the identifiability). 

Backward induction iterates through all decision points, starting from the last one $K$, where we define the Q-function and value function as follows:
\vspace{-3pt}
\begin{equation}\label{eq:K}\tag{1}
Q_K(\sigma_{K}, a_K) = \mathbb{E} \big[ y \mid \sigma_{K}, a_K \big]; \ \ \
V_K(\sigma_{K}) = \max_{a_K \in A_K} Q_K(\sigma_{K}, a_K).
\vspace{-6pt}
\end{equation}
If we then continue recursively, we define the $Q$-function and value function at decision point $k \in \{K-1, \ldots, 1 \}$ as
\vspace{-3pt}
\begin{equation}\label{eq:k}\tag{2}
Q_k(\sigma_{k}, a_k) = \mathbb{E} \big[ V_{k+1}(\sigma_{{k+1}}) \mid \sigma_{k}, a_k \big]; \ \ \
V_k(\sigma_{k}) = \max_{a_k \in A_k} Q_k(\sigma_{k}, a_k).
\vspace{-6pt}
\end{equation}
Then, under a perfectly known Q-function, the optimal actions and policy are given by
\vspace{-3pt}
\begin{equation}\label{eq:optimal}\tag{3}
\pi^{\text{opt}}_k(\sigma_{k}) = \arg\max_{a_k \in \mathcal{A}_k} Q_k(\sigma_{k}, a_k) = a^{opt}_k; \ \ \  \pi^{opt} = \{ \pi^{opt}_1(\sigma_{1}), \ldots, \pi^{opt}_K(\sigma_{K}) \}
\vspace{-6pt}
\end{equation}
which concludes the backward induction.

The value function can alternatively be expressed as: 
\vspace{-3pt}
\begin{equation}\label{eq:regret}\tag{4}
V_k(\sigma_{k}) = \mathbb{E} \big[ V_{k+1}(\sigma_{{k+1}}) + \big( Q_k(\sigma_{k}, a_k^{\text{opt}}) - Q_k(\sigma_{k}, a^{obs}_k) \big) \mid \sigma_{k} \big]
\vspace{-6pt}
\end{equation}
with $V_{K+1} = y$. This formulation, known as the \textit{regret-based} form, yields the same optimal policy.

Intuitively, backward induction works by starting at the last decision point $K$ and asking: `\textit{Given a current prefix $\sigma_{K}$, what is the expected outcome of each possible action?}'. This gives us the Q-function at $K$. The value function at $K$ then tells us the best outcome achievable from any $\sigma_{K}$ by selecting the highest Q-value (using the standard formulation at eq. \ref{eq:K} here for simplicity).  
We then move one step backward to decision point $K-1$. Here, the Q-function asks: `\textit{If I take action $a_{K-1}$ now for a prefix $\sigma_{{K-1}}$, and then act optimally at $K$ (with the new prefix $\sigma_{{K}}$ resulting from $a_{K-1}$), what outcome can I expect?}'. This is answered by the value function at $K$ for $\sigma_{{K}}$. The value function at $K-1$ again picks the action with the highest Q-value. 
We repeat this process all the way to the first decision point. In general, the Q-function answers: `\textit{How good is it to take a specific action now, assuming we act optimally afterward?}', while the value function answers: `\textit{How good is it to have this prefix, assuming optimal actions from here on?}'. By applying this procedure recursively, we can compute all Q-functions and thus determine the optimal policy.
% \vspace{-10pt}

\subsubsection{Causal learners.}
While backward induction provides the theoretically optimal solution, applying it in practice requires estimation. This is challenging because we rely on observational event logs, and, as discussed in Section \ref{sec:background}, observed confounders make estimation difficult. To address this, we draw inspiration from the DTR approaches in \cite{schulte2014,cclearning} and use causal learners to estimate the Q-function at each decision point and, consequently, the optimal action. While \textit{SCOPE} still uses function approximation to estimate Q-functions, it avoids explicit process approximations like MDP construction or data augmentation for generating training trajectories. Causal learners are designed to estimate outcomes and/or causal effects under different actions using observational data~\cite{metalearnersmulti}. These techniques try to explicitly reduce dependence on the historical decision policy (in contrast to~\cite{weinzierl2020}, see Section~\ref{sec:background}). 
Although \textit{SCOPE} supports any suitable causal learner, we apply three widely used, model-agnostic options compatible with any base model (e.g., an LSTM): the S-, T-, and RA-learner~\cite{metalearnersmulti}. We choose these for their stability, which is ideal for the PresPM setting. For example, unlike the IPW- and DR-learner, they avoid inverse propensity weights (IPW), which are used to handle confounding by reweighting inputs based on the inverse historical propensity of an action being taken for that input. However, in PresPM, the high input/prefix diversity (see Section~\ref{sec:background}) could produce near-zero probabilities, making IPWs highly unstable~\cite{metalearnersmulti}.
% We apply three widely adopted causal learners (the S-, T-, and RA-learner~\cite{metalearnersmulti}) that are model-agnostic: they can incorporate any base predictive model (e.g., an LSTM). We prefer these three as they align well with the PresPM setting: many other learners handle confounding via weighting by the historical probability of an action being taken for an input (e.g., IPW- and DR-learners), but in PresPM (see Section~\ref{sec:background}), the high diversity of inputs/prefixes can lead to near-zero probabilities and thus instability~\cite{metalearnersmulti}. Nonetheless, any suitable causal learner could be applied.
% In this paper, we apply three causal learners, widely adopted in CI: an S-learner, a T-learner, and an RA-learner~\cite{metalearnersmulti}, though any causal learner expected to perform well on a given dataset could be used. Each of these learners is model-agnostic: they can flexibly incorporate any suitable base predictive model (e.g., an LSTM).

We adopt the regret-based formulation~\ref{eq:regret} of the value function, which is more robust to model misspecification (e.g., incorrect model assumptions)~\cite{huang2015optimization}. Incorporating the relative difference (regret) between optimal and observed actions can reduce error propagation compared to estimating absolute values (as in the standard form~\ref{eq:k}). This is particularly useful in business processes, again because inputs (prefixes) can be highly heterogeneous due to variable control flow (see Section~\ref{sec:background}). Under this formulation, we must estimate two key components at each decision point $k$: the Q-function $Q_k$ and the corresponding optimal action $a_k^{opt}$. Once these are obtained, computing the value function becomes straightforward. Note that, while in the theoretical setting with a perfectly known Q-function, the optimal action is always given by the action that maximizes $Q_k$ (see eq. \ref{eq:optimal}), in practical approximation settings, the optimal action may be better estimated using alternative strategies to compensate for prediction error and model uncertainty, as we do below with the RA-learner. 

% \paragraph{S-learner.}The S-learner estimates $Q_k$ using a single predictive model~\cite{metalearnersmulti}. The model (e.g., an LSTM) takes as input the features in $\sigma_{k}$ together with the historical action taken $a^{obs}_k$. At inference time, we query the model once for every possible action $a_k$ at decision point $k$, producing estimates of $Q_k(\sigma_{k}, a_k)$. The chosen action is then simply the one with the highest predicted value.
\paragraph{S-learner.}The S-learner estimates $Q_k$ using a single model (e.g., an LSTM), taking as input the features in $\sigma_{k}$ together with the historical action taken $a^{obs}_k$~\cite{metalearnersmulti}. At inference time, the model is queried for each possible action $a_k$ at decision point $k$ to obtain an estimate of $Q_k(\sigma_{k}, a_k)$. The action with the highest estimate is selected.

% \paragraph{T-learner.}The T-learner builds a separate model for each possible action at decision point $k$~\cite{metalearnersmulti}. Each model takes as input only the features in $\sigma_{k}$. At inference, we query each model to obtain the predicted outcome under its corresponding action. These predictions serve as estimates of $Q_k(\sigma_{k}, a_k)$, and the best action is again the one with the largest predicted score.
\noindent\textit{T-learner.} The T-learner builds a separate model for each possible action at decision point $k$~\cite{metalearnersmulti}, using only features $\sigma_{k}$. At inference, each model estimates $Q_k(\sigma_{k}, a_k)$ under its corresponding action, and the action with the highest estimate is selected.

% \paragraph{RA-learner.}
% The RA-learner works in two steps~\cite{metalearnersmulti}. First, it estimates $Q_k$ in the same way as the S-learner. Then, it constructs a pseudo-outcome for each possible action. Intuitively, this pseudo-outcome represents the estimated causal effect of choosing action $a_k$ on the KPI, given the observed prefix. For a given prefix, observed action $a^{obs}_k$, decision point $k$, and action of interest $a_k$, the pseudo-outcome is:
\noindent\textit{RA-learner.}
The RA-learner works in two steps~\cite{metalearnersmulti}: it first estimates $Q_k$ like the S-learner, then constructs a pseudo-outcome for each possible action. For a prefix, observed action $a^{obs}_k$, decision point $k$, and action of interest $a_k$, the pseudo-outcome is:
\vspace{-5pt}
\begin{align*}\label{eq:po}\tag{5}
\Phi_k^{(a_k)}(\sigma_{k}) = \mathbf{1}\{a^{obs}_k = a_k\}(y - \hat{Q}(\sigma_{k},a_k)) + \sum_{f_k \neq a_k} \mathbf{1}\{a^{obs}_k = f_k\} (\hat{Q}(\sigma_{k},f_k) - y) 
\\ + \sum_{f_k \neq a_k} \mathbf{1}\{a^{obs}_k = f_k\}(\hat{Q}(\sigma_{k},f_k) - \hat{Q}(\sigma_{k},b_k)).\\[-20pt]
\end{align*}
% \vspace{-5pt}
where $f_k$ runs over all actions other than $a_k$, and $b_k$ is an arbitrary baseline action. These pseudo-outcomes (which still have the observed $y$ in eq. \ref{eq:po}, and thus cannot serve as estimation for new data) act as targets in a second predictive model, which outputs estimates of the causal effect of each action at decision point $k$. Intuitively, constructing pseudo-outcomes allows the second model to focus entirely on the causal effect (the difference in Q-values), which can be much simpler to learn than the Q-values (as in the other learners). The final action is chosen as the one with the largest predicted causal effect. The RA-learner is thus an explicit two-step approach for better causal effect estimation.
% \vspace{-20pt}
\vspace{-10pt}

\subsubsection{Algorithm.}
In summary, \textit{SCOPE} combines regret-based backward induction with causal learning to derive effective policies from observational event logs, handling cases with varying numbers of decision points and heterogeneous inputs (prefixes) resulting from variable control flow.
In Algorithm \ref{algo:S_full_case}, we provide an overview of the full training and inference procedure for an S-learner variant, referred to as \textit{SCOPE-S}. Lines highlighted in \textcolor{green!75!black}{green} mark the steps where the Q-function and optimal action at decision point $k$ are estimated. When adopting a T- or RA-learner, these steps are the ones that change most substantially, being replaced by the corresponding estimation procedures described above. Full algorithms for these other learners are also provided in our repository (see footnote \ref{fn:code}).

During training, the algorithm applies backward induction, starting from the last decision point. At each decision point, it uses a causal learner (here, an S-learner) to estimate the Q-function (line 4) and determine the optimal action for every case that reached that point (line 7). It then computes the estimated value for each relevant case using a regret-based formulation (lines 8–10). If a case did not visit the current decision point (due to variable control flow in business processes), the value from the subsequent decision point is carried backward (line 13). These values serve as targets for the preceding decision point.
During inference, the process moves forward. For each case, the algorithm observes the current prefix and checks whether a decision point is reached (lines 18–19). If so, it selects the model corresponding to that decision point to estimate and execute the optimal action (lines 20–21). This continues until the case is completed.
% \footnote{Algorithm~\ref{algo:S_full_case}: the action space can be predefined or discovered, e.g., as in~\cite{leoni2020}.}
% \vspace{-30pt}

\input{algorithms/S_full}
% \vspace{-10pt}

%% file: algorithms/S_full.tex
\begin{algorithm}

\footnotesize

\caption{SCOPE-S: Training and Inference}\label{algo:S_full_case}

\begin{algorithmic}[1]
    \Statex \textbf{Training (Offline)}
    \Statex \textit{Input:} Dataset $ \mathcal{D} = \bigcup_{c \in C} \bigcup_{k \in \mathcal{K}^{(c)}} \Big\{ \big(\sigma_{k}^{(c)}, a_k^{obs,(c)}, y^{(c)}\big) \Big\} $, action spaces $\{A_k\}_{k=1}^K$
    
    \State $\hat{V}_{K+1}^{(c)} = y^{(c)} \ \forall c \in \{1, \dots, C\}; \ \mathbf{M} = [\ ]$ \Comment{Initialize}
    % \State $\mathbf{M} = [\ ]$
    
    \For{$k = K, K-1, \dots, 1$} \Comment{Iterate backwards}
        
        \State $\mathcal{C}_k = \{c \mid k \in \mathcal{K}^{(c)}\}$ \Comment{Identify cases that visited $k$}
        
        \State \textcolor{green!75!black}{$\mathcal{M}_k \gets \text{Regress } \hat{V}_{k+1}^{(c)} \text{ on } (\sigma_{k}^{(c)}, a_k^{obs,(c)}) \text{ for all } c \in \mathcal{C}_k$} \Comment{S-learner Q-function}
        \State Append $\mathcal{M}_k$ to $\mathbf{M}$
        
        \For{each case $c \in \mathcal{C}_k$}
            \State \textcolor{green!75!black}{$\hat{a}^{opt,(c)}_k = \arg\max_{a_k \in A_k} \mathcal{M}_k(\sigma_{k}^{(c)}, a_k)$} \Comment{Estimated optimal action}
            
            \State $\hat{Q}_k^{obs,(c)} = \mathcal{M}_k(\sigma_{k}^{(c)}, a_k^{obs,(c)})$ \Comment{Q-value of observed action}
            \State $\hat{Q}_k^{opt,(c)} = \mathcal{M}_k(\sigma_{k}^{(c)}, \hat{a}^{opt,(c)}_k)$ \Comment{Q-value of optimal action}
            
            \State $\hat{V}_k^{(c)} = \hat{V}_{k+1}^{(c)} + \hat{Q}_k^{opt,(c)} - \hat{Q}_k^{obs,(c)}$ \Comment{Regret-based value update}
        \EndFor
        
        \For{each case $c \notin \mathcal{C}_k$}
            \State $\hat{V}_k^{(c)} = \hat{V}_{k+1}^{(c)}$ \Comment{Carry over value if point $k$ is skipped}
        \EndFor
    \EndFor
    
    \State \Return $\mathbf{M}$
    
    \vspace{0.3cm}
    
    \Statex \textbf{Inference (Online)}
    \Statex \textit{Input:} Models $\{\mathcal{M}_k\}_{k=1}^K$, initial prefix $\sigma_{k_{first}}^{(c)}$
    \While{case $c$ is not completed}
        \If{case $c$ reaches \textsc{decision point} $k \in \{1, \dots, K\}$}
            \State Observe current prefix $\sigma_{k}^{(c)}$ 
            \State \textcolor{green!75!black}{$\hat{a}^{opt,(c)}_k = \arg\max_{a_k \in A_k} \mathcal{M}_k(\sigma_{k}^{(c)}, a_k)$}
            \State Execute action $\hat{a}^{opt,(c)}_k$
        \EndIf
    \EndWhile

\end{algorithmic}
\end{algorithm}
\vspace{-7pt}

%% file: 4-experiments_discussion.tex
\section{Experiments \& Discussion}\label{sec:experiments_discussion}
\vspace{-5pt}
This section outlines the (semi-)synthetic data, methods for comparison, the experimental procedure, and finally the results and implications\footnote{\label{fn:code}The code, algorithms, simulators, and additional results are available at \url{https://github.com/JakobDeMoorKULstudent/SCOPE}}.
% \vspace{-10pt}
\vspace{-10pt}

\subsection{Data}
\vspace{-3pt}
Synthetic and semi-synthetic data are widely used in causal machine learning because real-world datasets do not provide ground-truth counterfactuals~\cite{holland1986}. In sequential PresPM, this challenge is even stronger, since for each case we only observe the KPI under the historically executed intervention sequence, not under the many alternative sequences that could have been recommended. Real-life event logs therefore do not allow direct ground-truth evaluation of learned sequential intervention policies.
%Given an offline dataset, we cannot always directly verify whether the model’s recommendation would have led to a better outcome for a test case. We only observe what actually happened under the historical decision policy, not what would have happened under a different intervention action~\cite{holland1986}.
To our knowledge, SimBank is the only simulator designed specifically for PresPM, which we use in our evaluation~\cite{SimBank}. We also introduce a new semi-synthetic setup based on the BPIC17\_W event log~\cite{bpic17} and release it publicly to support further research in sequential PresPM.

\vspace{-10pt}

\subsubsection{SimBank.}
SimBank models a bank’s loan application process and has been thoroughly validated. The main KPI to optimize is the total profit generated from loans in a test set. Among other options, we opt for the following two sequential interventions for this research:
\begin{enumerate*}
    \item \textit{Choose procedure}: decide between a standard or priority procedure
    \item \textit{Set interest rate}: choose one of 3 possible interest rate levels.
\end{enumerate*}
The effects of these interventions vary across clients, making the decisions complex. The two decision points are highly interdependent. Selecting a standard procedure reduces costs, but increases the likelihood of a client refusal of the offer in the end. This refusal probability directly influences the optimal choice of interest rate, which must balance affordability for the client (to reduce refusals) against profitability (to cover costs, which are higher for priority procedures). The first decision thus affects both client refusal and costs, which are critical considerations in the second decision. Note that historically, the second decision point is sometimes (in minority of cases) skipped entirely due to the bank's policy. SimBank allows us to vary the level of confounding in the training data by adjusting the \% of confounded data (generated fully under the bank's policy) versus full RCT data (generated by employing a random policy in the decision points). As shown in previous studies, higher confounding typically reduces the performance of PresPM methods~\cite{bozorgi2023CI,SimBank}.
\vspace{-10pt}

\subsubsection{SimBPIC17}
We additionally introduce SimBPIC17, a semi-synthetic simulation based on the BPIC17\_W event log, which records real-world loan application cases. In this simulation, we simplify the control-flow sligthly (by deleting 3 activities) and introduce an intervention on the existing activity `call\_incomplete\_files'. At each decision point, the choice is whether to \textit{call} the client to follow up on incomplete files or to \textit{wait} for the client to provide the information themselves. The KPI to minimize is the total cost incurred while completing the files: $cost_{files} = cost_{tpt}*tpt_{files} + n\_calls*cost_{call}$, where $tpt_{files}$ is the throughput time of completing the files, which is sampled from a uniform distribution, and $cost_{tpt}$ and $cost_{call}$ are fixed constants. There is a trade-off: waiting is cheaper but increases throughput time, whereas calling is more expensive but shortens the throughput time. The (causal) effect of a call on the throughput time depends on two factors: the loan type (with ‘car’ and ‘loan takeover’ having larger effects) and the average duration of all previous activities (longer durations increase the potential time saved by calling). This dependency on average duration creates strong inter-decision point interactions: a call in decision point 1 reduces throughput time but also lowers the average duration of activities after decision point 1, which in turn reduces the effect of a call in decision point 2.
To generate datasets, variables except activity labels and activity durations are sampled from the original dataset. We then define the control-flow, the function determining the effect of a call, the function calculating the KPI based on the actions taken, and the bank policy, which governs whether the bank chooses to call or wait. The bank calls if the loan type is ‘car’ or ‘loan takeover’ and the average activity duration exceeds 4025 units. Unlike SimBank, SimBPIC17 makes it easy to vary the number of decision points by increasing the number of (obligatory) choices between call or wait. We mimic an observational event log by defining the bank policy, varying confounding levels in the same way as in SimBank.
% Similar to SimBank, defining the bank policy allows us to mimic an observational event log, and we vary confounding levels by adjusting the probability of choosing a random action versus following the bank policy; this is equivalent to how we vary confounding in SimBank.
\vspace{-10pt}

\subsection{Methods for comparison}
\vspace{-5pt}
We compare \textit{SCOPE} with the multi-interventional approaches of Branchi et al.~\cite{branchi2022} and de Leoni et al.~\cite{leoni2020}, each missing a property from Table~\ref{tab:positioning}: ~\cite{branchi2022} uses a process approximation, while ~\cite{leoni2020} does not align interventions across decision points. Including both methods allows us to examine the role of these properties, which \textit{SCOPE} explicitly includes.

We select Branchi et al.~\cite{branchi2022} as the representative method using a process approximation and exclude Abbasi et al.’s FORLAPS~\cite{abbasi2025} (see Section \ref{sec:background}). Although FORLAPS supports sequential interventions through process approximation, it is inflexible for adapting to our optimization problems. It defines MDP states solely by observed activity labels, making comparison unfair because our interventions require additional process variables. Incorporating these variables would require expanding the state space (producing an impractically large Q-table), redesigning the state representation, or introducing new augmentation methods—changes that would fundamentally alter the original method. In contrast, Branchi et al. approximate the process as an MDP and train a Q-learning algorithm using KMeans clustering, where each state is represented by the last executed activity and the cluster label assigned to the current prefix. While the original work defines states using only control-flow information, the method can be easily extended to include additional variables by incorporating them into the clustering inputs, which is how we apply it. We refer to this approach as \textit{KMeans-Q}.

As our second baseline, the method of de Leoni et al.~\cite{leoni2020} represents an approach that does not align interventions across decision points. The original method trains a single model and selects actions by predicting the KPI for each possible action at every decision point, equivalent to using a single S-learner for all decision points. We adapt this approach by using a separate model per decision point, equivalent to applying an S-learner (separately) at each step. This modification is theoretically equivalent to the original method and we implement it for three reasons: (1) it simplifies the implementation of causal learners when decision variables vary across points; (2) despite this complexity of implementation, we initially implemented~\cite{leoni2020} as a single model, but it yielded worse results; and (3) it thus ensures a fairer comparison with \textit{SCOPE}, which also uses a model per decision point. Because these models are applied independently, the action chosen at one decision point does not account for the impact of future actions. We refer to this approach as SEP-S, indicating the use of separate S-learners. More generally, \textit{SEP} refers to any strategy in our experiments that trains separate models for each decision point.

We also include a simple \textit{random policy} as a baseline, and an \textit{upper bound} representing the best achievable KPI, computed by taking, for each test case, the maximum KPI obtainable across all possible action sequences (exhaustive search).

\vspace{-10pt}
% \vspace{-15pt}

\subsection{Preprocessing, Hyperparameter tuning \& Dataset splits}
\vspace{-8pt}
Cases are truncated at decision points to form prefixes ending before the intervention. 
LSTMs use tensor encoding \cite{tensorencoding}, with case-level features added at the final layer. Other models use last-event encoding for time features and aggregation encoding for other features, following \cite{teinemaaoutcome}. Categorical features are one-hot encoded and continuous features standardized. Depending on the learner, an action feature may be included (S- or RA-learner) or not (T-learner).
For KMeans-Q, we adopt the original preprocessing~\cite{branchi2022} for control-flow features and add non–control-flow features using last-event and aggregation encoding as before.
% During preprocessing, cases are truncated at the decision points, producing prefixes that end immediately before the intervention. LSTMs use tensor encoding~\cite{tensorencoding} with full event sequences and case-level features incorporated at the final layer. Other models use last-event encoding for time features and aggregation encoding for remaining features (because of its proven effectivenss~\cite{teinemaa2018}). Categorical variables are one-hot encoded and continuous features standardized. Depending on the learner, an action feature may be included (S- or RA-learner) or not (T-learner).
% During preprocessing, cases are truncated at predefined decision points, producing prefixes that end immediately before the intervention. As described later, we employ different model types. For LSTMs, prefixes are transformed using tensor encoding~\cite{tensorencoding}, with full event sequences processed and case-level features incorporated at the final layer. Models that do not support tensor encoding use last-event encoding for time features and aggregation encoding for all other features, given its proven effectiveness in outcome-oriented predictive process monitoring~\cite{teinemaa2018}. Categorical variables are one-hot encoded, while continuous features are standardized. Depending on the learner, we may also include an action feature (for an S-learner) or not (for a T-learner).

Hyperparameters are tuned via Bayesian optimization using 20\% of the training set for validation. Models are then retrained on the full training set. For evaluation, we use 10,000 test cases for SimBank and 1,000 for SimBPIC17 (due to the exponential growth of sequences with decision points). Both \textit{SCOPE} and SEP can use any base model, tuned identically for fair comparison. In KMeans-Q, we jointly tune the KMeans and RL components using a normalized metric that combines silhouette score and average reward, allowing twice the tuning budget compared to \textit{SCOPE} and SEP due to the dual-model setup. The original paper tuned KMeans first using clustering metrics (e.g., silhouette score) and then tuned the RL model. In our experiments, this approach performed poorly, as optimizing KMeans for clustering quality alone does not support effective intervention decisions.
% \footnote{The original paper tuned KMeans first using clustering metrics (e.g., silhouette score) and then tuned the RL model. In our experiments, this approach performed poorly, as optimizing KMeans for clustering quality alone does not support effective intervention decisions.}
% For every method and dataset, we tune the hyperparameters using Bayesian optimization. We use 20\% of the training set for validation. After selecting the best hyperparameters, each method is retrained on the full training set. For the evaluation phase, we use 10,000 test cases for the SimBank simulations, while for SimBPIC17 we use 1,000 test cases. The smaller size for SimBPIC17 is due the computational cost of varying the number of decision points, which results in an exponential growth in the number of possible sequences for a case ($2^{n\_{\text{decisionpoints}}}$).
% Both \textit{SCOPE} and SEP can use any base model, and we tune these base models in the same way to ensure a fair comparison. For KMeans-Q, we need to tune both the KMeans clustering model and the RL component. In their original paper, the authors first tuned the KMeans model using a clustering evaluation metric (e.g., silhouette score), and then tuned the RL model afterward. However, in our experiments we found that this approach leads to poor performance, because the KMeans model is optimized for clustering quality rather than for supporting effective intervention decisions. Therefore, we jointly tune the KMeans and RL models using silhouette score and average reward in a single normalized metric.
\vspace{-10pt}

\subsection{Experimental Setup}
\vspace{-7pt}
To evaluate \textit{SCOPE}, we use the \textit{Gain}, which measures the \% improvement in total KPI achieved by a method’s policy over the bank’s historical decision policy on the test set. We vary key simulation parameters relevant to sequential PresPM, and do this especially in SimBank, as it is thoroughly validated in~\cite{SimBank}, contains more variables, and has a more complex causal structure than SimBPIC17 (where we focus on the number of decision points).
In SimBank, we adjust:
\begin{itemize}[noitemsep, topsep=0pt, leftmargin=*]
    \item \textit{Confounding level $\delta$ (0.9--0.99):} Reflects real-world observational data, which is usually heavily confounded (as there is generally already a decision policy in place in business processes). Varying this helps assess whether a method is robust to biases introduced by the historical decision policy (e.g., bank policy) and whether it can still select effective actions even when those actions are rarely observed in the data.
    \item \textit{Training set size (1K, 10K, 50K):} Since each model at a decision point in \textit{SCOPE} depends on the previous one, limited data may cause estimation errors that propagate through decision points. Varying training size helps measure this effect.
    \item \textit{Learners and base models:} We test different learners (S-, T-, RA-learner) and base models (XGBoost, Random Forest, LSTM/MLP) for \textit{SCOPE} and SEP, to examine whether \textit{SCOPE}’s advantages (due to its aligned interventions) hold consistently across learner types and underlying base model. The default is S-learner + XGBoost (e.g., for varying training size).
\end{itemize}
\vspace{3pt}

In SimBPIC17, the focus is evaluating performance over more than 2 decision points. We vary the number of decision points from 2 to 6 and run experiments under 3 levels of confounding (0.9, 0.95, 0.99) to assess robustness across longer decision horizons. 
Each setting is run using 10 different random seeds in models. 

In our repository (see footnote~\ref{fn:code}), we provide: (1) results across different learners and base models for SimBPIC17 as well, confirming consistency across datasets; and (2) a computational complexity analysis showing that, although \textit{SCOPE} has the highest theoretical training cost, it matches SEP and outperforms the original implementation of~\cite{leoni2020} in inference complexity. While KMeans-Q is theoretically the most efficient, \textit{SCOPE} trained significantly faster in practice, had only slightly slower inference than SEP simply due to obtaining more complex models from tuning, and can even outperform KMeans-Q at inference time.
\vspace{-10pt}
% \vspace{-8pt}

\subsection{Results}
\vspace{-6pt}
Figure \ref{fig:training_sizes} shows method performance across confounding levels and \textbf{training sizes} on SimBank. \textit{SCOPE} and SEP are implemented as S-learners with XGBoost (SEP thus corresponds to de Leoni et al.’s SEP-S). \textit{SCOPE} outperforms nearly all settings, except for the (1K, 0.99) case, demonstrating its strong ability to learn sequential intervention policies. Performance generally declines with higher confounding, except for KMeans-Q, and in the 50K condition, where having more data keeps performance high even for strong confounding. As expected, larger training sets improve performance (for KMeans-Q, this is expressed in less variability). \textit{SCOPE}’s advantage grows with training size, as each model uses the outputs of other models in later decision points as targets through the value function. With more data, model errors decrease and propagate less across decision points. KMeans-Q performs the worst and most variable, suggesting that approximating the process for RL while still aligning all interventions can be less effective than treating interventions independently, as SEP(-S) does.
% Figure \ref{fig:training_sizes} presents the performance of each method across confounding levels and \textbf{training sizes} on SimBank, where \textit{SCOPE} and SEP are implemented as S-learners with XGBoost as the base model for fair comparison, meaning SEP corresponds to the method of de Leoni et al.(SEP-S). \textit{SCOPE} performs best in nearly all settings, with the single exception of the (1K, 0.99) case, demonstrating its strong ability to learn effective sequential intervention policies. As confounding increases, performance generally declines, except for KMeans-Q at the 1K size, and in the 50K training-size condition, where all methods benefit from the large amount of data. As expected, all methods improve with larger training sets, but the performance difference of \textit{SCOPE} over other methods grows as training size increases. This is because during training of \textit{SCOPE}, a model uses the outputs of other models in later decision points as targets through the value function. With more data, model errors decrease and propagate less across decision points. KMeans-Q performs the worst, suggesting that approximating the process to train an RL agent while still aligning all interventions can sometimes be less effective than treating interventions independently, as SEP(-S) does.

\input{figures/figure_training_sizes}

Figure \ref{fig:learners} compares \textit{SCOPE} and SEP across confounding levels and \textbf{learner types} on SimBank (10K training cases, XGBoost). \textit{SCOPE} consistently outperforms SEP, with only occasional exceptions at the 0.99 confounding level. Overall, the S-learner and RA-learner achieve the strongest performance, while the T-learner shows the largest degradation under high confounding. Similarly, Figure \ref{fig:base_models} shows performance across confounding levels and \textbf{base models} (10K training cases, S-learner)\footnote{For the right-most plot, we use an MLP for the early decision point and and an LSTM in the later decision point, where sequential structure becomes important.}. \textit{SCOPE-S} again outperforms SEP-S. XGBoost achieves the best overall performance, while Random Forest performs worst. Both these plots suggest aligning interventions across decision points is generally more effective than treating them independently, regardless of learner or base model type.
% Figure \ref{fig:learners} compares \textit{SCOPE} and SEP across confounding levels and \textbf{learner types} on SimBank (10K training cases, XGBoost). \textit{SCOPE} consistently outperforms SEP, except for the (T-learner, 0.99) setting, indicating that aligning interventions across decision points is generally more effective than treating them independently. Overall, the S-learner performs best, the T-learner worst, and the RA-learner’s performance declines least under high confounding, consistent with prior CI findings~\cite{curthNonParam, kunzel2017}.
% Figure \ref{fig:learners} shows the performance of \textit{SCOPE} and SEP across confounding levels and \textbf{learner types} on SimBank, using 10K training cases and XGBoost as the base model. \textit{SCOPE} consistently performs best, with the sole exception of the (T-learner, 0.99) setting. This indicates that aligning interventions across decision points is generally more effective than treating them as independent decisions, regardless of the learner used. Overall, the S-learner achieves the strongest performance on SimBank, while the T-learner performs the worst. At very high confounding levels, the RA-learner’s performance declines the least relative to its performance under lower confounding, aligning with prior findings in the CI literature~\cite{curthNonParam, kunzel2017}.

Figure \ref{fig:numbers_of_decision_points} shows method performance across different \textbf{numbers of decision points} and confounding levels on SimBPIC17 (10K training cases, \textit{SCOPE} and SEP are again S-learners with XGBoost). \textit{SCOPE} consistently performs best, and its advantage over other methods increases as the number of decision points grows because it better captures interdependent decisions. Both \textit{SCOPE} and SEP-S gain more over the bank policy with more decision points but drift further from the upper bound—\textit{SCOPE} due to error accumulation across models, and SEP due to independent treatment of decision points. KMeans-Q degrades with more decision points, likely because the KMeans model increasingly aggregates data relative to its original form, reducing the quality of the MDP approximation for Q-learning.
\input{figures/figure_learners}
\input{figures/figure_base_models}
\input{figures/figure_numbers_of_decision_points}
\vspace{-15pt}

\subsection{Discussion}
% In Summary, the setup of \textit{SCOPE} demonstrates clear advantages across a range of settings. Its performance improves substantially with larger training sets, as it reduces any accumulation of errors in the backward induction. This benefit also increases with more decision points, since \textit{SCOPE} can effectively align interdependent interventions, whereas methods that treat decision points independently leave those dependencies unexploited. The method consistently outperforms separate optimization regardless of learner type or base model. To fully leverage backward induction from observational data, practitioners should select the learner–base-model combination they expect to perform best. S-learners are sufficient for our datasets, but RA-learners seem more robust under strong confounding, and T-learners struggle. While RL-based approaches that approximate the entire process as an MDP aim for joint optimization, in practice they can perform worse than simply treating decision points independently, as we observe with KMeans-Q. \textit{SCOPE} does rely on the predictive accuracy of underlying models: a highly unpredictable KPI can lead to strong error accumulation. However, his challenge affects all methods, and \textit{SCOPE} still likely retains its advantage by leveraging observational event logs directly and considering the interdependent structure of decision points.
\vspace{-5pt}
In summary, the setup of \textit{SCOPE} demonstrates clear advantages across a range of settings. Its performance improves with larger training sets by reducing error accumulation in backward induction and scales well with more decision points by aligning interdependent interventions, something separate optimization fails to exploit.  \textit{SCOPE} consistently outperforms separate optimization across learner types and base models. Practitioners should choose the learner–model combination expected to perform best: S- and RA-learners suffice for our datasets, but T-learners struggle. While RL-based approaches that approximate the entire process as an MDP (like KMeans-Q) aim for joint optimization, they can still perform worse than treating decision points independently (as in SEP). \textit{SCOPE} does rely on the predictive accuracy of underlying models: a highly unpredictable KPI can lead to strong error accumulation. However, this challenge affects all methods, and \textit{SCOPE} still likely retains its advantage by leveraging observational event logs directly using causal learners and considering the interdependent decision point structure.
% \vspace{-10pt}
\vspace{-5pt}

\paragraph{Limitations.}
\textit{SCOPE} models sequential dependencies within individual process cases but assumes no interference between interventions across different cases, as required by the SUTVA assumption (Section \ref{sec:methodology}). Similarly, sequential ignorability may not always hold in business processes. Despite these assumptions—common to most PresPM approaches—our experiments show that \textit{SCOPE} can still identify effective sequential policies under realistic levels of confounding typical in business processes.
Moreover, \textit{SCOPE} theoretically increases training complexity by requiring a separate predictive model and backward induction at each decision point. However, as shown in our repository (see footnote~\ref{fn:code}), our practical results differ: despite being theoretically most demanding, \textit{SCOPE} trained significantly faster than KMeans-Q. During inference, \textit{SCOPE} matches SEP’s theoretical complexity (and is more efficient than the original implementation of~\cite{leoni2020}), with only slightly longer runtimes in practice due to obtaining more complex tree models during hyperparameter tuning.
% \paragraph{Limitations.}\textit{SCOPE} accounts for sequential dependencies within each process case, but assumes there is no interference between interventions across different process cases. This assumption, part of the SUTVA assumption described in Section \ref{sec:methodology}, is also made by all other sequential PresPM approaches.
\vspace{-10pt}

%% file: figures/figure_training_sizes.tex
\vspace{-5pt}
\begin{figure}[!ht]
    \centering
    \includegraphics[width=1\textwidth]{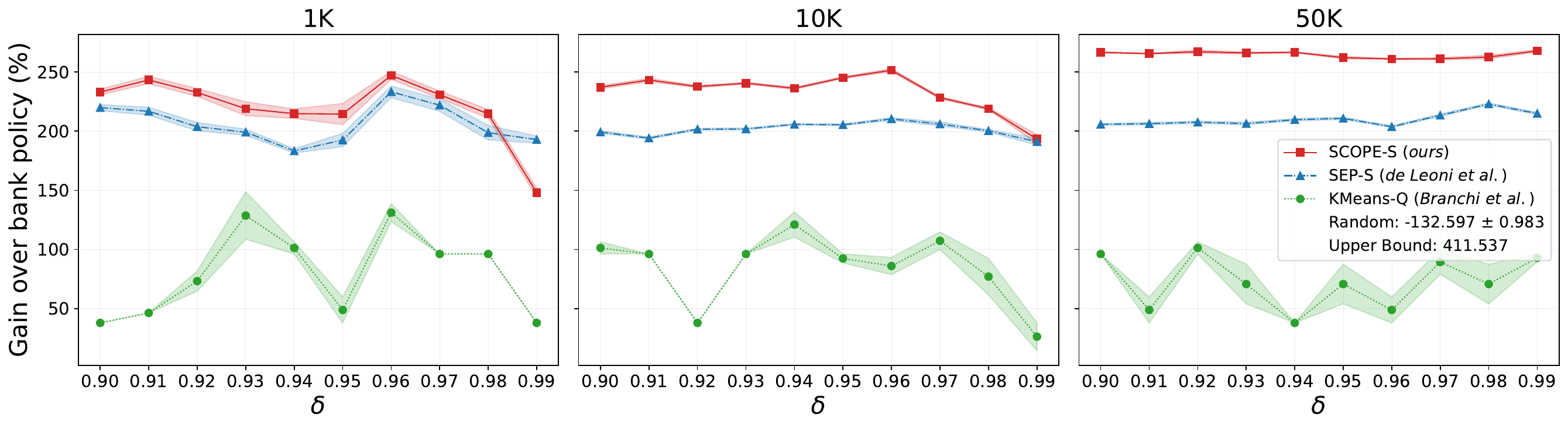}
    \caption{Gain on SimBank across confounding levels ($\delta$) for \underline{different training sizes}. The shaded area shows one standard error over 10 iterations. \textit{SCOPE} and Sep: S-learner with XGBoost.}
    \label{fig:training_sizes}
\end{figure}
\vspace{-2pt}

%% file: figures/figure_learners.tex
\begin{figure}[!ht]
    \centering
    \includegraphics[width=1\textwidth]{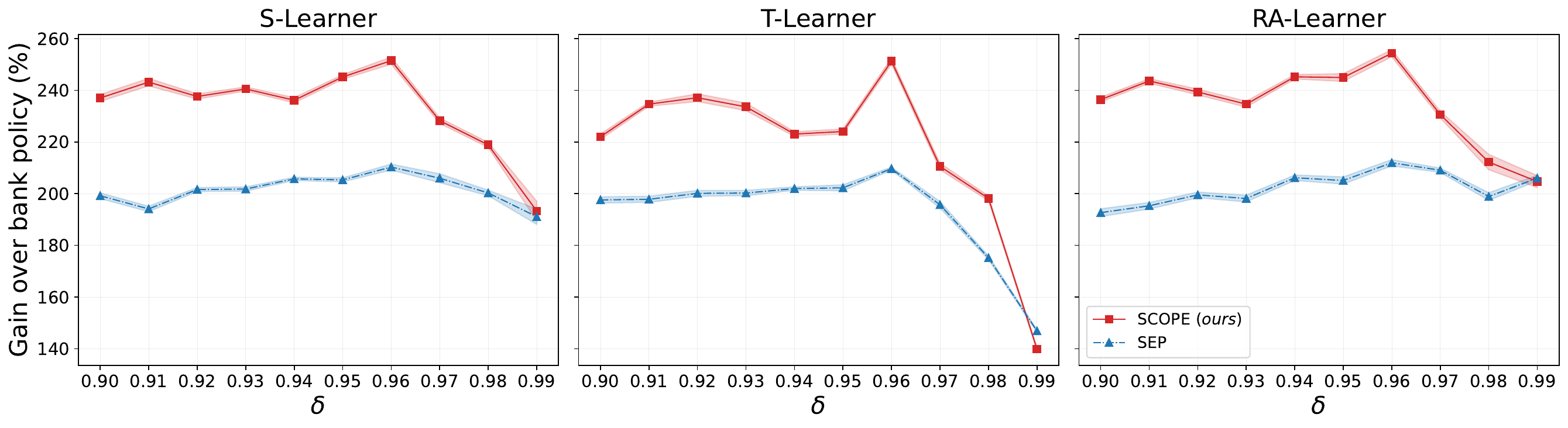}
    \caption{Gain on SimBank across confounding levels ($\delta$) for \underline{different learners}. The shaded area shows one standard error over 10 iterations. \textit{SCOPE} and Sep: XGBoost, trained on 10K cases.}
    \label{fig:learners}
% \vspace{-10pt}
\end{figure}

%% file: figures/figure_base_models.tex
\begin{figure}[!ht]
    \centering
    \includegraphics[width=1\textwidth]{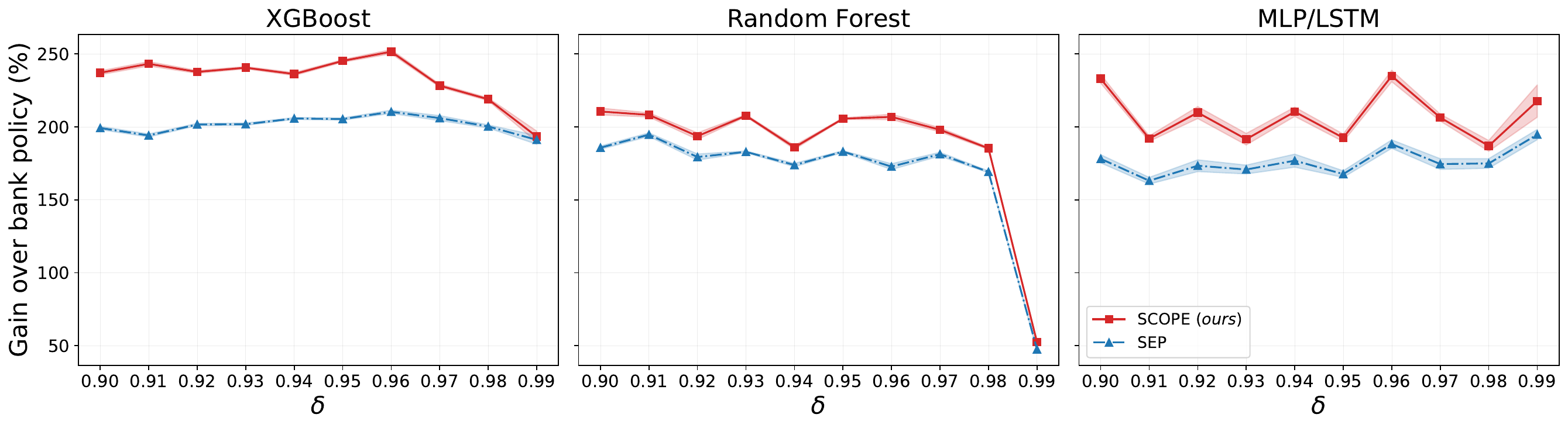}
    \caption{Gain on SimBank across confounding levels ($\delta$) for \underline{different base models}. The shaded area shows one standard error over 10 iterations. \textit{SCOPE} and Sep: S-learner trained on 10K cases.}
    \label{fig:base_models}
% \vspace{-10pt}
\end{figure}

%% file: figures/figure_numbers_of_decision_points.tex
\begin{figure}[!ht]
    \centering
    \includegraphics[width=1\textwidth]{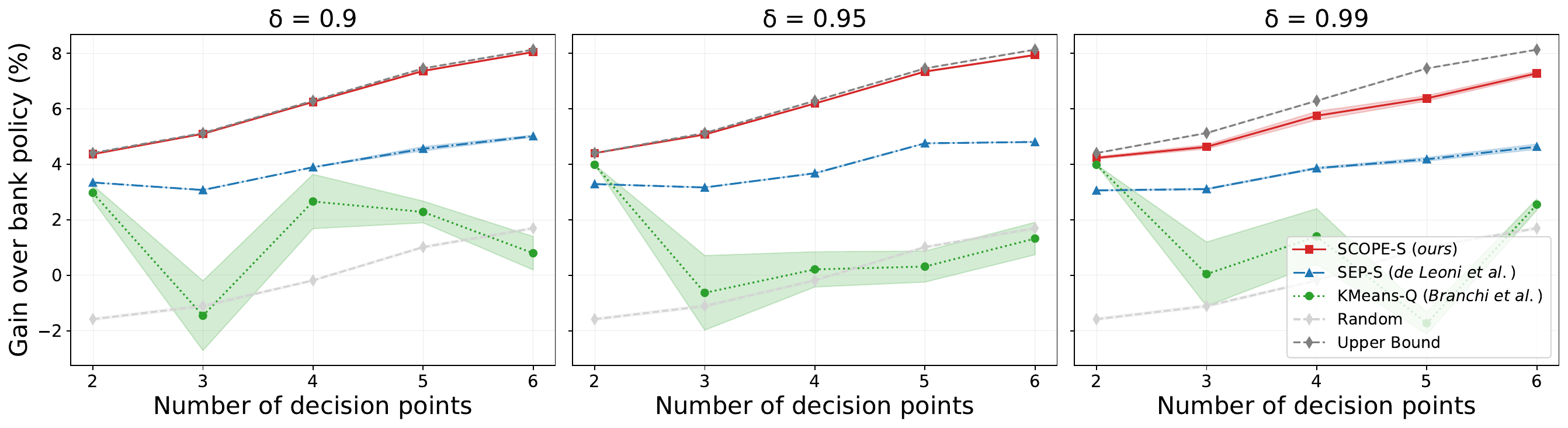}
    \caption{Gain on SimBPIC17 across \underline{varying numbers of decision points} for three confounding levels ($\delta$), trained on 10K cases. The shaded area shows one standard error over 10 iterations. \textit{SCOPE} and Sep: S-learner with XGBoost.}
    \label{fig:numbers_of_decision_points}
\end{figure}

%% file: 5-conclusion.tex
\section{Conclusion}\label{sec:conclusion}
\vspace{-10pt}
We introduce \textit{SCOPE}, a method that combines backward induction with causal learning to optimize sequential interventions in business processes. Using a series of experiments, including a newly developed semi-synthetic simulator to support further research in sequential PresPM, we show that \textit{SCOPE} outperforms existing sequential PresPM methods, which either handle each intervention independently or rely on approximating the process to train an RL agent.

Future work could extend SCOPE to account for interference (e.g., via causal methods for interference~\cite{caljon}) and relax sequential ignorability (e.g., using instrumental variables~\cite{ivCI}). Computational cost could potentially be reduced via an end-to-end neural framework (e.g.,~\cite{igcnet}) adapted to backward induction.
% Future work could explore ways to extend \textit{SCOPE} to account for interference between cases while still utilizing backward induction and causal learning. For instance, this could involve investigating causal methods that explicitly handle the presence of interference~\cite{caljon,netest}.
\vspace{-10pt}